%% file: main.tex
\documentclass[letterpaper, 10 pt, conference]{ieeeconf}

\input{preamble}

\title{\LARGE \bf Learning Humanoid Arm Motion via Centroidal Momentum Regularized Multi-Agent Reinforcement Learning}

\author{Ho Jae Lee$^{1}$, Se Hwan Jeon$^{1}$, and Sangbae Kim$^{1}$%
\thanks{$^{1}$Department of Mechanical Engineering, Massachusetts Institute of Technology, Cambridge, MA 02139, USA
{\tt\small \{hjlee201, sehwan, sangbae\}@mit.edu}}%
}

\AtBeginBibliography{\footnotesize}

\begin{document}

\maketitle
\thispagestyle{empty}
\pagestyle{empty}

\allowdisplaybreaks
\input{commands}

\input{Sections/00_abstract}
\input{Sections/01_intro}
\input{Sections/02_background}
\input{Sections/03_cm_computation}

\input{Sections/04_marl}

\input{Sections/05_results}
\input{Sections/06_conclusion}

\addtolength{\textheight}{0cm}
\footnotesize{\printbibliography}
\end{document}

%% file: preamble.tex
\IEEEoverridecommandlockouts             
\usepackage{graphics} 
\usepackage{epsfig} 
\usepackage{times} 
\usepackage{amsmath} 
\usepackage{amssymb}  
\usepackage[per-mode=symbol]{siunitx}
\usepackage{hyperref}
\usepackage{algorithm}
\usepackage{algpseudocode}
\usepackage{listings}
\usepackage[font=footnotesize]{caption}
\usepackage{subcaption}
\usepackage[isbn=false,
    backend=biber,
    style=ieee,
    bibstyle=ieee,  
    sortcites=true,   
    giveninits=true,            
    backref=false]{biblatex}
\usepackage{xcolor}
\usepackage{color,soul}
\definecolor{codegreen}{rgb}{0,0.6,0}
\definecolor{codegray}{rgb}{0.5,0.5,0.5}
\definecolor{codepurple}{rgb}{0.58,0,0.82}
\definecolor{backcolour}{rgb}{0.95,0.95,0.92}

\lstdefinestyle{mystyle}{
  backgroundcolor=\color{backcolour}, commentstyle=\color{codegreen},
  keywordstyle=\color{magenta},
  numberstyle=\tiny\color{codegray},
  stringstyle=\color{codepurple},
  basicstyle=\ttfamily\scriptsize,
  breakatwhitespace=false,         
  breaklines=true,                 
  captionpos=b,                    
  keepspaces=true,                 
  showspaces=false,                
  showstringspaces=false,
  showtabs=false,                  
  tabsize=2
}
\let\Re\relax
\DeclareMathOperator{\Re}{\mathbb{R}}

\DeclareMathOperator{\xvec}{\mathbf{x}}

%% file: commands.tex
\newcommand{\algorithmautorefname}{Algorithm}
\renewcommand{\figureautorefname}{Fig.}
\renewcommand{\equationautorefname}{Eq.}
\renewcommand{\sectionautorefname}{Section}
\renewcommand{\subsectionautorefname}{Section}

\newcommand{\casadi}{\texttt{casadi}}
\newcommand{\cusadi}{\texttt{CusADi}}

\newcommand{\CoMPos}{\mathbf{r}}                        
\newcommand{\CoMAcc}{\ddot{\mathbf{r}}}                 

\newcommand{\CM}{\mathbf{h}_G}              
\newcommand{\CLM}{\mathbf{l}_G}             
\newcommand{\CAM}{\mathbf{k}_G}             
\newcommand{\dCM}{\dot{\mathbf{h}}_G}       
\newcommand{\dCLM}{\dot{\mathbf{l}}_G}      
\newcommand{\dCAM}{\dot{\mathbf{k}}_G}      
\newcommand{\CMBase}{\mathbf{h}_{G,\text{base}}}              
\newcommand{\CMLegs}{\mathbf{h}_{G,\text{legs}}}              
\newcommand{\CMArms}{\mathbf{h}_{G,\text{arms}}}              

\newcommand{\CMM}{\mathbf{A}_G}              
\newcommand{\dCMM}{\dot{\mathbf{A}}_G}       
\newcommand{\genPos}{\mathbf{q}}             
\newcommand{\genVel}{\dot{\mathbf{q}}}       
\newcommand{\genAcc}{\ddot{\mathbf{q}}}      

\newcommand{\armObs}{\mathcal{O}_{\text{arm}}}                 
\newcommand{\armJointPos}{\mathbf{q}_{\text{arm}}}             
\newcommand{\armJointVel}{\dot{\mathbf{q}}_{\text{arm}}}       
\newcommand{\prevArmAction}{\mathbf{a}_{\text{arm}}^{t-1}}     
\newcommand{\legObs}{\mathcal{O}_{\text{leg}}}                 
\newcommand{\legJointPos}{\mathbf{q}_{\text{leg}}}             
\newcommand{\legJointVel}{\dot{\mathbf{q}}_{\text{leg}}}       
\newcommand{\prevLegAction}{\mathbf{a}_{\text{leg}}^{t-1}}     
\newcommand{\phase}{\phi}
\newcommand{\baseLinVel}{\boldsymbol{v}_\base}                 
\newcommand{\baseAngVel}{\boldsymbol{\omega}_\base}            

\newcommand{\armAction}{\mathcal{A}_{\text{arm}}}                 
\newcommand{\armJointPosDes}{\Delta\hat{\boldsymbol{q}}_{\text{arm}}}  
\newcommand{\legAction}{\mathcal{A}_{\text{leg}}}                 
\newcommand{\legJointPosDes}{\Delta\hat{\boldsymbol{q}}_{\text{leg}}}  
\newcommand{\refJointPos}{\boldsymbol{q}^\text{ref}}              

\newcommand{\CMDes}{\hat{\mathbf{h}}_G}              
\newcommand{\CAMDes}{\hat{\mathbf{k}}_G}              
\newcommand{\CLMDes}{\hat{\mathbf{l}}_G}              
\newcommand{\genVelDes}{\hat{\dot{\mathbf{q}}}}       
\newcommand{\zeroVec}{\textbf{0}}       

\newcommand{\base}{\mathcal{B}}             
\newcommand{\velcmd}{\hat{\boldsymbol{v}}_\base}        
\newcommand{\velcmdX}{\hat{v}_\base^x}                  
\newcommand{\velcmdY}{\hat{v}_\base^y}                  
\newcommand{\velcmdZ}{\hat{\omega}_\base^z}             
\newcommand{\linVelBase}{\boldsymbol{v}_\base}             
\newcommand{\angVelBase}{\boldsymbol{\omega}_\base}             
\newcommand{\velBase}{\Tilde{\boldsymbol{v}}_\base}             

\newcommand{\velBaseX}{v_\base^x}                       
\newcommand{\velBaseY}{v_\base^y}                       
\newcommand{\velBaseZ}{\omega_\base^z}                  
\newcommand{\contactSchedule}{\phi_{\text{contact}}}    

\newcommand{\policy}{\boldsymbol{\pi}}      
\newcommand{\gradient}{\hat{g}}             
\newcommand{\advantage}{\hat{A}}            

\newcommand{\verticalGRM}{\boldsymbol{M}^z}            
\newcommand{\FixedArms}{\textit{Fixed Arms}}
\newcommand{\WithArmsWoCAM}{\textit{With Arms w/o CAM}}
\newcommand{\WithArms}{\textit{With Arms}}
\newcommand{\SingleAgent}{\textit{Single Agent}}
\newcommand{\MultiAgentDTDE}{\textit{Multi-Agent DTDE}}
\newcommand{\MultiAgentCTCE}{\textit{Multi-Agent CTCE}}

\newcommand{\comment}[1]{{\color{blue}Comment: #1}}
\newcommand{\hojae}[1]{{\color{cyan}Ho Jae: #1}}
\newcommand{\sehwan}[1]{{\color{orange}Se Hwan: #1}}
\newcommand{\mithumanoid}{a humanoid platform}%

\newcommand{\todo}[1]{{\color{orange}TODO: #1}}
\newcommand{\fixme}[1]{{\color{red}FIXME: #1}}
\newcommand{\needref}{{\color{blue}[REF]}}

\newcommand{\ie}{i\/.\/e\/.,\/~}%
\newcommand{\eg}{e\/.\/g\/.,\/~}%
\newcommand{\cf}{cf\/.\/~}%

%% file: Sections/00_abstract.tex
\begin{abstract}

Humans naturally swing their arms during locomotion to regulate whole-body dynamics, reduce angular momentum, and help maintain balance. 
Inspired by this principle, we present a limb-level multi-agent reinforcement learning (RL) framework that enables coordinated whole-body control of humanoid robots through emergent arm motion. 
Our approach employs separate actor-critic structures for the arms and legs, trained with centralized critics but decentralized actors that share only base states and centroidal angular momentum (CAM) observations, allowing each agent to specialize in task-relevant behaviors through modular reward design.
The arm agent guided by CAM tracking and damping rewards promotes arm motions that reduce overall angular momentum and vertical ground reaction moments, contributing to improved balance during locomotion or under external perturbations.
Comparative studies with single-agent and alternative multi-agent baselines further validate the effectiveness of our approach.
Finally, we deploy the learned policy on \mithumanoid, achieving robust performance across diverse locomotion tasks, including flat-ground walking, rough terrain traversal, and stair climbing.
\end{abstract}

%% file: Sections/01_intro.tex
\section{INTRODUCTION}
Arm swing is a natural and characteristic feature of human locomotion, but its fundamental role remains unclear.
Biomechanical studies have offered a variety of explanations and consequences of arm swing during locomotion.
Some argue that it reduces the induced angular momentum of the legs, ensuring smoother movement~\cite{elftman1939function}.
Consequently, the required ground reaction moments for locomotion are lower~\cite{herr2008angular}, improving energy economy of gait patterns~\cite{collins2009dynamic}.
For humanoid platforms, however, it is less clear how to effectively and intentionally coordinate the arms during locomotion, especially with typical optimization or learning-based control methods.
Designing such controllers is difficult due to the complexities of whole-body dynamics, the variability of locomotion tasks, and uncertainty on how arm motion should be guided with respect to the task, stylistically or otherwise.

Prior work has typically approached controlling arm and leg motion in a unified manner.
Imitation-based approaches try to match distributions of reference data, as in \cite{peng2022ase,tang2024humanmimic,peng2017deeploco}.
These controllers are able to exhibit human-like behavior, but it is unclear how (or if) mimicking these references offers advantages for locomotion—the primary reward signal is to imitate trajectories from higher degree-of-freedom systems, which is not necessarily aligned with robustness or performance for the specific platform.
With reinforcement learning (RL) methods, potentially conflicting rewards make it challenging to plan arm and leg motions simultaneously, and arm motion is often regularized with heuristics that may have little physical basis other than stylistic tuning~\cite{radosavovic2024real}.
This balance becomes even more sensitive with loco-manipulation, such as in \cite{sleiman2023versatile}, where different sets of limbs can be responsible for entirely different tasks.

Model-based controllers have typically leveraged \textit{full-body} optimization schemes to generate arm motion.
Via centroidal momentum tracking~\cite{kajita2003resolved, orin2013centroidal, wensing2016improved}, relative angular-acceleration control~\cite{miyata2019walking}, or full-body model predictive control (MPC)~\cite{dai2014whole,khazoom2022humanoid} arm motion can arise naturally with a physically-consistent basis.
However, solving these full-body motion plans becomes difficult to scale for online, closed-loop evaluation, often requiring approximations or special treatment for sufficient bandwidth.
Additional considerations, such as avoiding self-collisions, state estimation noise, and environmental constraints, can further complicate these model-based controllers.
While reasoning about the full dynamics of the legs and arms is ideal, the computational expense can make these control schemes intractable.

\begin{figure}[t]
    \centering
    \includegraphics[width=0.9\linewidth]{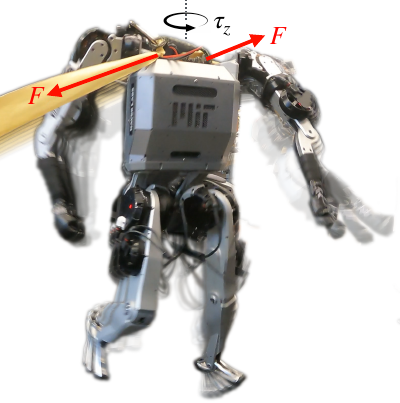}
    \caption{
    Coordinated arm response under external torque disturbance.
    Two external forces $F$ are applied from opposite sides to induce a positive yaw torque disturbance $\tau_z$ about the vertical axis. The robot reacts with emergent arm motions that counteract the induced momentum, helping to recover and maintain balance.}
    \label{fig:HW_push}
    \vspace{-3mm}
\end{figure}

In contrast to these whole-body approaches, there is evidence that the arms and legs are \textit{not} planned jointly at the biological level.
Commands for the arms and legs arise from entirely separate cortical areas, and are only coupled for stabilization elsewhere in the cerebellum~\cite{graziano2007mapping,dietz2001neuronal}.
It is unclear whether this is simply vestigial.
However, this separation in control can present several advantages from an engineering perspective.
Optimizations for complex systems can be greatly accelerated by "sectioning" the physics of the model, such as the front and rear legs of a quadruped, and ensuring consensus downstream~\cite{amatucci2024accelerating}.
Similarly, locomotion can be trained separately from manipulation tasks by abstracting arm motion as random inertial wrenches on the body, as in~\cite{ma2022combining}.
This controller design enables the limbs to be reasoned about separately without mutual interference, allowing for coordinated motion without intractable computation or meticulous tuning.


Inspired by this architecture, we approach this challenge of coordinating locomotion with arm motion as a \textit{ multi-agent RL problem}—the arms and legs are treated as separate agents in the learning environment, connected only by the centroidal dynamics of the robot.
Specifically, we adopt a centralized training and decentralized execution (CTDE) approach \cite{lowe2017multi, foerster2018counterfactual}, wherein decentralized actors for the arms and legs observe only their respective proprioceptive information, while centralized critics have access to global state information from both limb groups.
This architecture enables faster convergence during training to a higher asymptotic task reward, and we analyze the effect of having centralized/decentralized observations for both the actor and critic.

Additionally, we introduce a centroidal angular momentum reward to guide arm motion, motivated by prior biomechanical studies of locomotion and physically-grounded principles.
The resultant controller demonstrates natural arm swing while walking and improved recovery from external disturbances, which we validate on hardware for \mithumanoid.
We analyze the ground reaction moments and momentum during execution and find they corroborate the findings from~\cite{elftman1939function,herr2008angular,collins2009dynamic}.

We summarize our main contributions as follows:
\begin{enumerate}
\item We introduce a CAM reward based on biomechanical studies of human walking, and find that it guides the emergence of natural arm swing for stable locomotion and effective push recovery for our policy.
\item We propose a multi-agent RL framework employing separate actor-critic networks for the arms and legs, trained centrally but executed in a decentralized manner.
\item We demonstrate the effectiveness and practicality of our controller by validating its performance on \mithumanoid, both in simulation and hardware experiments\footnote{\href{https://youtu.be/uPC4OlujM3k}{Youtube video}: https://youtu.be/uPC4OlujM3k}.
\end{enumerate}

%% file: Sections/02_background.tex
\section{Background}

\begin{figure*}[tb]
    \centering
    \includegraphics[width=0.9\textwidth]{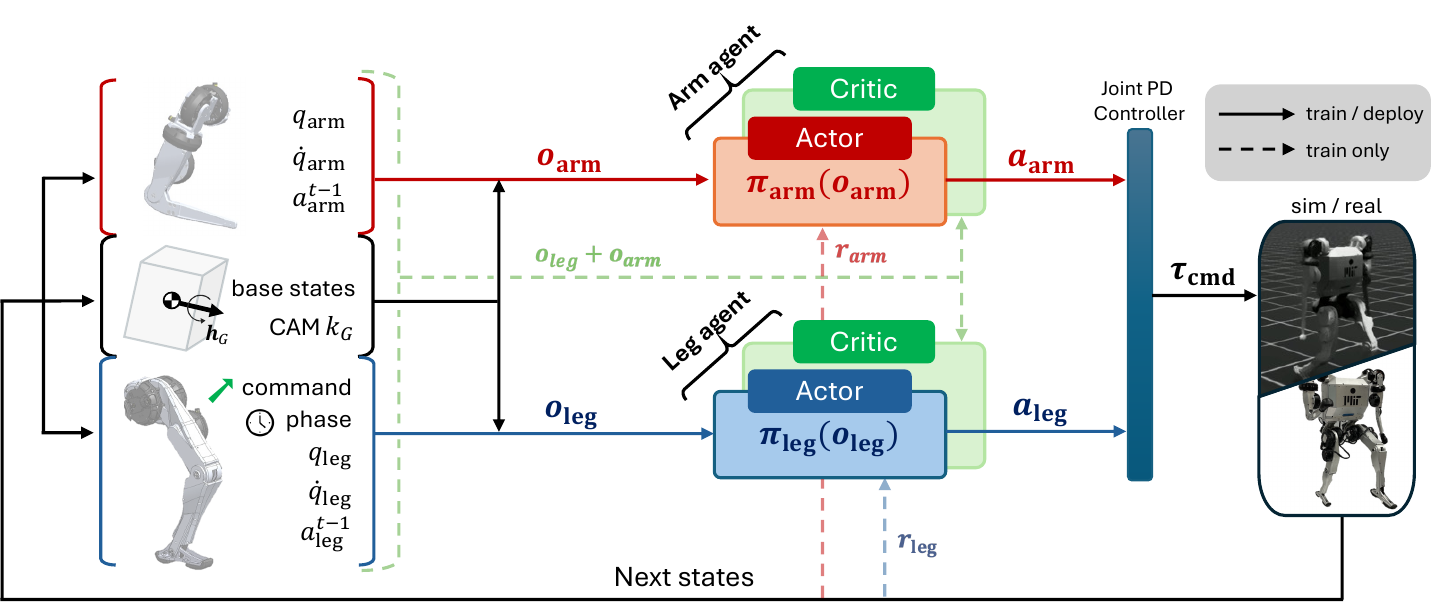}
    \caption{Overview of our limb-level multi-agent reinforcement learning framework with CAM regularization. Separate actor-critic policies are assigned to the arm and leg agents, each operating on distinct observation spaces and reward structures. During training, each critic is provided global information for centralized training, while actors execute using local observations, sharing only base states and CAM. This design enables coordinated whole-body behavior, as the arm and leg agents are implicitly coupled through the shared CAM signal, allowing the arm to respond to momentum generated by leg motion.}
    \label{fig:overview}
    \vspace{-6mm}
\end{figure*}

\subsection{Centroidal Dynamics and Momentum}
\label{section:centroidal_dynamics}

The centroidal dynamics of a system \cite{orin2013centroidal, wensing2016improved, dai2014whole} offers a balanced representation of robot dynamics, bridging the gap between simplified point-mass models and computationally expensive full-body dynamics.
It is particularly valuable in humanoid control, as it captures whole-body interactions in a compact form, enabling emergent upper-body motion and improving robustness within a whole-body control framework.
Formally, centroidal dynamics describes the linear and angular momentum dynamics around the robot’s center of mass (CoM) as follows:
\begin{eqnarray}
    \label{eq:centroidal_linear_dynamics}
    \dCLM = m\CoMAcc = \sum_{j} \mathbf{F_j} + m\mathbf{g}  \\
    \label{eq:centroidal_angular_dynamics}
    \dCAM = \sum_{j} (\mathbf{c_j} - \CoMPos) \times \mathbf{F_j} + \mathbf{\tau_j}
\end{eqnarray}
where $m$ denotes the total mass of the robot, $\CoMPos \in \Re^3$ is the CoM position, $\mathbf{F}_j \in \Re^3$ and $\boldsymbol{\tau}_j \in \Re^3$ are the external force and moment at contact point $\mathbf{c}_j \in \Re^3$, respectively, and $\mathbf{g} \in \Re^3$ represents gravitational acceleration.
Additionally, $\CLM \in \Re^3$ denotes the centroidal linear momentum (CLM), $\CAM \in \Re^3$ denotes the centroidal angular momentum (CAM), and the subscript $G$ indicates quantities expressed with respect to the CoM frame.

Centroidal momentum $\CM$, comprising both CAM and CLM, is expressed compactly as:
\begin{eqnarray}
\label{eq:centroidal_momentum}
    \CM(\genPos, \genVel) = 
    \begin{bmatrix} 
        \CAM \\ 
        \CLM
    \end{bmatrix}
        = \CMM(\genPos) \genVel 
\end{eqnarray}
where $\CMM \in \Re^{6 \times (n+6)}$ is the centroidal momentum matrix (CMM), $\genPos \in \Re^{(n+7)}$ and $\genVel \in \Re^{(n+6)}$ denote the generalized positions and velocities of the robot, respectively.

The generalized velocity vector $\genVel$ can be partitioned as:
\begin{equation}
\label{eq:gen_vel}
\genVel = \begin{bmatrix} \angVelBase^\top & \linVelBase^\top & \genVel_{\text{legs}}^\top & \genVel_{\text{arms}}^\top \end{bmatrix}^\top
\end{equation}
where $\angVelBase \in \Re^3$, $\linVelBase \in \Re^3$ represent the body angular and linear velocities in base frame, respectively, and $\genVel_{\text{legs}} \in \Re^{n_{\text{legs}}}$ and $\genVel_{\text{arms}} \in \Re^{n_{\text{arms}}}$ are the joint velocities of legs and arms, respectively.

By time-differentiating Eq \eqref{eq:centroidal_momentum}, we obtain the time-rate of change of centroidal momentum as:
\begin{eqnarray}
\label{eq:diff_centroidal_momentum}
    \dCM(\genPos, \genVel) = 
    \begin{bmatrix} 
        \dCAM \\ 
        \dCLM
    \end{bmatrix}
        = \CMM(\genPos) \genAcc +  \dCMM(\genPos) \genVel
\end{eqnarray}

Since external disturbances directly affect centroidal momentum, we exploit this relationship for reward shaping: specifically, we incentivize arm policies to actively reduce the rate of change of CAM, thereby enhancing overall locomotion stability.

\subsection{Multi-Agent Reinforcement Learning}
\label{section:marl}

Traditional policy gradient methods \cite{lillicrap2015continuous, schulman2017proximal} are poorly suited to multi-agent environments due to the inherent non-stationarity induced by simultaneously learning agents. 
As each agent's policy evolves during training, the environment appears non-stationary from the perspective of any individual agent, which undermines the convergence of standard RL algorithms.
To address this, the Centralized Training with Decentralized Execution (CTDE) paradigm \cite{lowe2017multi, foerster2018counterfactual} was introduced.
Under CTDE, each agent maintains its own actor that operates solely on local observations, while the corresponding critic is trained with access to additional global information, such as the actions and observations of other agents.
This centralized critic formulation stabilizes training by making the environment stationary with respect to each agent’s critic, even as policies change.

Formally, for any policy $\policy_i \neq \policy_i'$, the transition probability satisfies $P(s'|s, a_1, \cdots, a_N, \policy_1, \cdots, \policy_N) = P(s'|s, a_1, \cdots, a_N) = P(s'|s, a_1, \cdots, a_N, \policy_1', \cdots, \policy_N')$ where $s$ denotes the underlying environment state and $a_i$ denotes the action taken by agent $i$.
In the general multi-agent policy gradient formulation, the estimate of the policy gradient $\gradient$ based on the advantage function takes the form:
\begin{equation}
\label{eq:policy_gradient}
\gradient = \hat{\mathbb{E}}_t [\nabla_{\theta_i}\text{log}\policy_i(a_i|o_i)\advantage_i^t(\xvec, a_1, \cdots, a_N)]
\end{equation}
where $\advantage_i^t$ is the advantage estimate for agent $i$ computed using the centralized critic.
Here, $\xvec$ denotes a global state representation (typically a concatenation of all agents' observations) and $o_i$ represents the local observation available to agent $i$.

This framework is also compatible with any actor-critic algorithm.
A variant of Proximal Policy Optimization (PPO) \cite{schulman2017proximal} known as Multi-Agent PPO (MAPPO), has shown strong empirical performance in multi-agent settings \cite{yu2022surprising}.

%% file: Sections/03_cm_computation.tex
\section{Centroidal Momentum Modeling and Computation}
\label{section:CAM_computation}

\begin{figure}[t]
    \centering
    \includegraphics[width=0.98\linewidth]{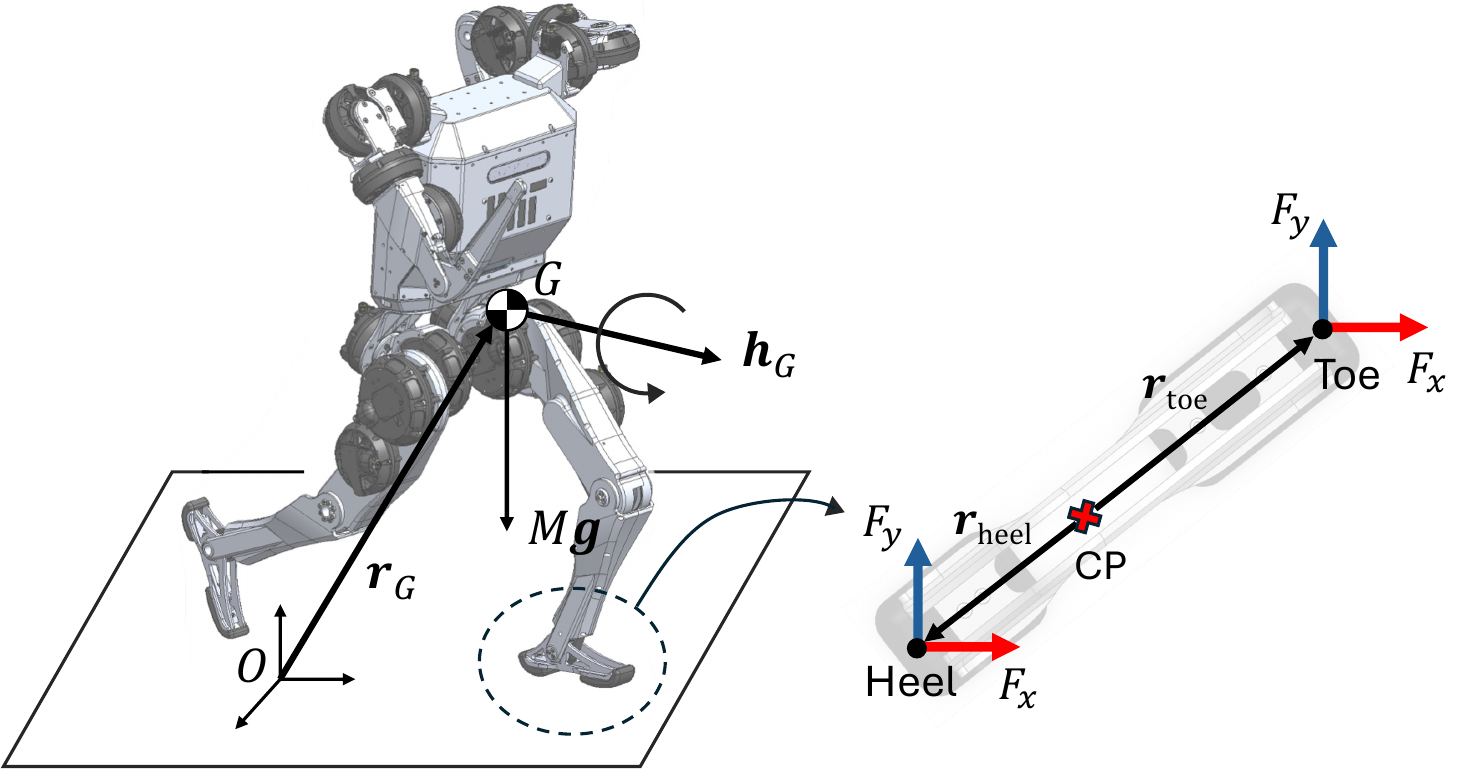}
    \caption{
    \textit{Left:} Illustration of \mithumanoid~with indicated total centroidal momentum $\CM$ at the center of mass ($G$).
    \textit{Right:} Top-down view of the foot with the cylindrical collision primitive, which permits only two contact points at the toe and heel. The corresponding horizontal contact forces $F_x$, $F_y$ are depicted at each point. 
    The vertical ground reaction moment (GRM) is then computed around the stance foot's center of pressure (CP).}
    \label{fig:CM_CP}
    \vspace{-4mm}
\end{figure}

Our experimental platform is \mithumanoid~\cite{saloutos2023design} (\autoref{fig:CM_CP}), which weighs 24.89 kg in total, with the arms accounting for 5.71 kg—approximately 22.8\% of the robot's mass.
The robot has 10 actuated joints for the legs and 8 for the arms.
To compute dynamic quantities such as the CMM and its derivatives, we use \texttt{Pinocchio} \cite{carpentier2019pinocchio}, a modern rigid-body dynamics library that provides efficient algorithms for computing forward dynamics, kinematics, and momentum mappings. 
We implement a \casadi \ \cite{andersson2019casadi} symbolic version of Pinocchio for analytical expression generation, and parallelize these expressions using \cusadi \ \cite{jeon2024cusadi}, enabling high-throughput batched evaluation across multiple simulation environments.

We compute the total centroidal momentum and its time derivative using \autoref{eq:centroidal_momentum} and \autoref{eq:diff_centroidal_momentum}.
To obtain the limb-wise contributions, we decompose $\CM = \CMBase + \CMLegs + \CMArms$ into additive components from the base, legs, and arms. 
This is achieved by multiplying the relevant block of the CMM with the corresponding segment of the generalized velocity vector, effectively isolating each limb's contribution to the total centroidal momentum.

We also define a reference centroidal momentum $\CMDes$ by projecting the desired base linear and angular velocities through the CMM evaluated at the current generalized position $\genPos$:
\begin{equation}
\label{eq:CM_des}
\CMDes(\genPos, \genVelDes) = 
\begin{bmatrix} 
    \CAMDes \\ 
    \CLMDes
\end{bmatrix}
= \CMM(\genPos) \genVelDes
\end{equation}
where 
$\genVelDes = 
\begin{bmatrix}
0 & 0 & \hat{\omega}_\base^z & \hat{v}_\base^x & \hat{v}_\base^y & 0 & \zeroVec_{\text{leg}}^\top & \zeroVec_{\text{arm}}^\top
\end{bmatrix}^\top$ denotes the reference generalized velocity embedding the user command $\velcmd = (\velcmdX, \velcmdY, \velcmdZ)$,  corresponding to the desired base linear velocity in the $x$ and $y$ directions, and angular velocity around the $z$-axis.
This reference serves as a target momentum profile in our reward shaping, encouraging the emergence of natural arm motions that contribute to coordinated whole-body locomotion.

%% file: Sections/04_marl.tex
\section{Limb-Level Multi-Agent Reinforcement Learning Framework}
\label{section:marl}


Taking inspiration from the separation of control present in the cortex, we approach learning whole-body control as a multi-agent RL problem, and train policies that separately control the arms and legs of the humanoid system (\autoref{fig:overview}).
By assigning distinct agents for the legs and arms of the platform, we can tailor observations and rewards for these subsystems individually.
This ensures that policy gradient updates from the leg agent do not interfere with the arm agent, and vice versa.

Each agent is trained using the PPO algorithm \cite{schulman2017proximal} in the IsaacLab simulation environment \cite{mittal2023orbit}, leveraging input normalization and parallel rollouts across 4096 environments.
Both arm and leg policies operate at a control frequency of 100 Hz, and their outputs are added to nominal joint positions $\refJointPos\in\mathbb{R}^{18}$, which are then passed to a low-level joint PD controller running at 1 kHz to compute motor torques.

\subsection{Arm agent}
\label{section:arm_agent}

The primary role of the arms in this work is to act as momentum generators, reacting to adjust the CAM to desired quantities. 
The observation space of the arm policy, denoted as $\armObs\in\mathbb{R}^{41}$, includes both arm-specific and base-related information. 
It consists of the base height, base heading, base linear and angular velocities in the base frame, the projection of the normalized gravity vector in the base frame, arm joint positions $\armJointPos$ and velocities $\armJointVel$, the previous arm action $\prevArmAction$, as well as the current and target CAM—where the $x$ and $y$ components are expressed in the base frame, and the $z$ component is expressed in the world frame.
The critic of the arm agent receives a global observation of size $\mathbb{R}^{76}$, which includes the full state of both the arm and leg joints, all base information, and both agents’ actions, enabling centralized training.
The action space $\armAction\in\mathbb{R}^{8}$ corresponds to the residual arm joint position commands $\armJointPosDes$.

We design the arm policy’s reward structure to encourage arm movements that regulate CAM, consisting of two primary components: a vertical CAM tracking term (\autoref{eq:CAM_tracking_reward}) and a horizontal CAM damping term (\autoref{eq:CAM_damping_reward}). 
The tracking reward $r_{\text{CAM}}$ encourages anti-phase arm swing relative to leg motion by aligning the vertical CAM with a reference CAM (\autoref{eq:CM_des}), while the damping reward $r_{\text{dCAM}}$ penalizes undesirable build-up of horizontal angular momentum, thereby promoting arm responses that dissipate perturbation-induced momentum. 
In addition, we include the arm regularization terms that follow the same structure as \cite{lee2024integrating}.
\begin{equation}
\label{eq:CAM_tracking_reward}
r_{\text{CAM}} = \exp(-(\frac{\CAMDes^z - \CAM^z}{1 + \lvert\CAMDes^z\rvert})^2/\sigma)
\end{equation}

\begin{equation}
\label{eq:CAM_damping_reward}
r_{\text{dCAM}} = -\min\left(0, \sum_{i \in \{x, y\}} \CAM^i \cdot \dCAM^i \right)
\end{equation}

\begin{figure}[t]
    \centering
    \includegraphics[width=0.95\linewidth]{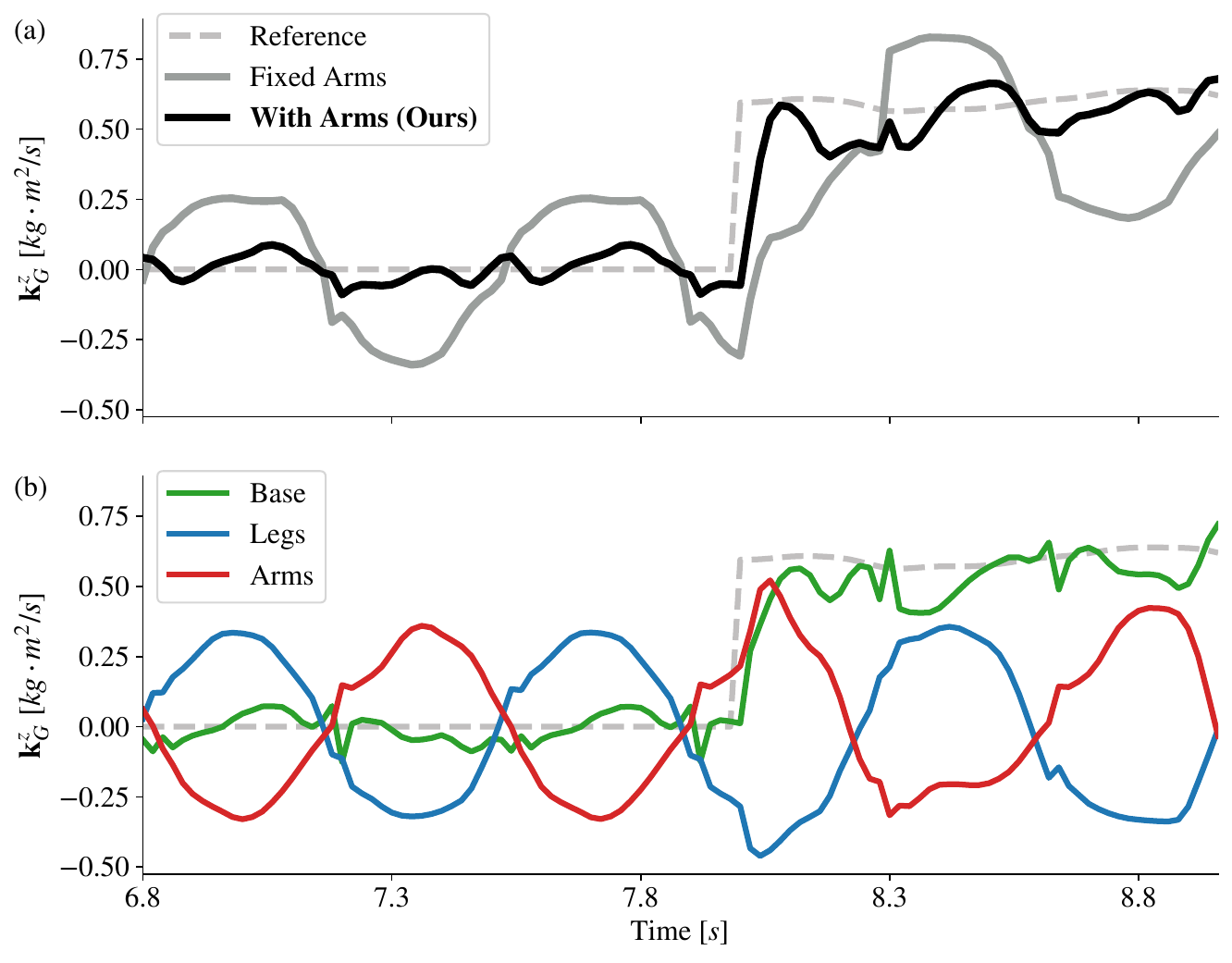}
    \caption{(a) Comparison of total vertical CAM between our method ($\WithArms$), $\FixedArms$ baseline, and the reference. Our method is able to track the desired CAM more closely by leveraging momentum from the arms. 
    (b) Decomposition of the total vertical CAM of our method into contributions from the base, legs, and arms. The angular momentum from the arms and legs effectively cancels each other out, resulting in improved tracking of the reference vertical CAM.}
    \label{fig:CAM}
    \vspace{-3mm}
\end{figure}

\subsection{Leg agent}
\label{section:leg_agent}

The primary objective of the leg agent is to track user-specified base velocity commands while adhering to the desired contact frequency.
The observation space of the leg policy, denoted as $\legObs\in\mathbb{R}^{49}$, includes leg-specific and base-related information.
It shares the same observations of the base and  CAM as the arm policy, in addition to the velocity command $\velcmd$, the contact phase signal $\phi$, leg joint positions $\legJointPos$, velocities $\legJointVel$, and previous leg action $\prevLegAction$.
Its critic shares the same global observations as the arm agent critic.
The action space $\legAction\in\mathbb{R}^{10}$ represents the residual leg joint position commands $\legJointPosDes$.

We formulate the leg policy’s reward to encourage accurate tracking of the user-defined base velocity commands.
It is comprised of two primary components: a velocity tracking reward (\autoref{eq:velocity_reward}) and a contact schedule reward (\autoref{eq:contact_reward}).
The velocity tracking term $r_{vt}$ encourages the base link to follow the commanded velocity, while the contact schedule term $r_{cs}$ incentivizes adherence to the specified stepping frequency.
\begin{equation}
\label{eq:velocity_reward}
r_{vt} = \exp(-(\frac{\velcmd - \velBase}{1 + \lvert\velcmd\rvert})^2/\sigma)
\end{equation}

\begin{equation}
\label{eq:contact_reward}
r_{cs} = (\mathbb{I}_{\text{r,contact}} - \mathbb{I}_{\text{l,contact}}) \contactSchedule
\end{equation}
Here, $\velcmd$ denotes the commanded base velocity defined in \autoref{section:CAM_computation}, while $\velBase = (\velBaseX, \velBaseY, \velBaseZ)$ represents the current base linear velocity in the $x$ and $y$ directions, and angular velocity around the $z$-axis, expressed in the base frame.
The terms $\mathbb{I}_{\text{r,contact}}$, $\mathbb{I}_{\text{l,contact}}$, $\phi$, and $\contactSchedule$ follow the definitions introduced in \cite{lee2024integrating}.
As with the arm policy, we also include additional leg regularization terms to ensure smooth motions.

%% file: Sections/05_results.tex
\section{Results}
\label{sec:results}

\begin{figure}[t]
    \centering
    \includegraphics[width=0.95\linewidth]{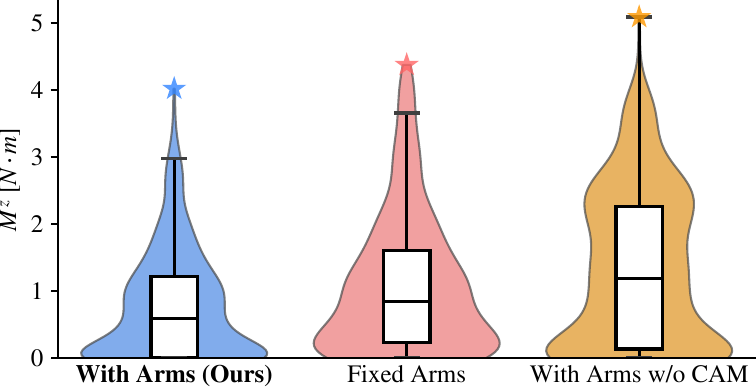}
    \caption{Comparison of vertical ground reaction moment (GRM) $\verticalGRM$ distributions across three controllers: our proposed method, $\FixedArms$, and $\WithArmsWoCAM$, evaluated over 10 walking steps. Our method exhibits a narrower GRM distribution and the lowest peak moment, aligning with the conclusions from~\cite{collins2009dynamic}. Notably, the passive arm-swing policy ($\WithArmsWoCAM$) results in higher GRM peaks and a wider distribution, suggesting that uncontrolled arm motion can negatively impact locomotion stability.}
    \label{fig:GRM_violin}
    \vspace{-3mm}
\end{figure}

In this section, we investigate the benefits of our architecture and reward shaping.
We find that leveraging CAM-based rewards induces coordinated and naturalistic arm motion, improving the system's ability to recover from disturbances. 
Analyses of the ground reaction moments from the policy reflect those from Collins et al., suggesting that arm swing can reduce undesirable wrenches on the body~\cite{collins2009dynamic}.
Moreover, our limb-level multi-agent RL framework improves learning stability and overall task performance compared to standard single-agent learning.
Our results are evaluated on \mithumanoid~in both simulation and hardware experiments.







\subsection{CAM-regularized arm motion}
\label{sec:cam_arm_motion}
In this section, we investigate the emergent arm behavior induced by CAM-based reward shaping and analyze its contribution to the improved whole-body locomotion.
To highlight the effect of CAM regularization, we compare our methods with two baselines:
\begin{itemize}
  \item \textbf{Fixed Arms}: A baseline where only the legs are actuated and arms remain fixed.
  \item \textbf{With Arms w/o CAM}: A full-body policy that actuates both arms and legs, using the same architecture and training setup as our method, but excluding CAM-related rewards (\autoref{eq:CAM_tracking_reward}, \autoref{eq:CAM_damping_reward}).
\end{itemize}

As shown in \autoref{fig:CAM}, our method tracks the reference CAM far more accurately compared to the $\FixedArms$ baseline.
The decomposition of the vertical CAM contributions from the base, legs, and arms shows that the momentum from the arms and legs is consistently out of phase, effectively canceling each other and resulting in better tracking of the reference vertical CAM, consistent with the findings from~\cite{collins2009dynamic}.

But why is minimizing fluctuations in vertical CAM crucial? 
Increased fluctuations in vertical CAM can be attributed to larger contact wrenches (\autoref{eq:centroidal_linear_dynamics}, \autoref{eq:centroidal_angular_dynamics}), expressed as increased vertical ground reaction moment (GRM) around the stance foot's center of pressure (CP) as depicted in \autoref{fig:CM_CP}.
\autoref{fig:GRM_violin} compares the distribution of vertical GRM $\verticalGRM$ for our method against $\FixedArms$ and $\WithArmsWoCAM$ over 10 walking steps. 
Our method shows a narrower GRM distribution and achieves the smallest peak vertical moment, consistent with observations in \cite{collins2009dynamic}. 
This result highlights that properly controlled arm motion guided by CAM observations effectively reduces vertical GRM, thereby decreasing foot slippage and improving overall locomotion stability.
Interestingly, the $\WithArmsWoCAM$ policy performs the worst, exhibiting both the highest peak vertical moment and consistently elevated GRMs throughout the gait cycle. 
This result suggests that passively swinging arms without intentional CAM-based control may introduce greater instability compared to intentionally fixed arms during locomotion.

\begin{figure}[t]
    \centering
    \includegraphics[width=0.98\linewidth]{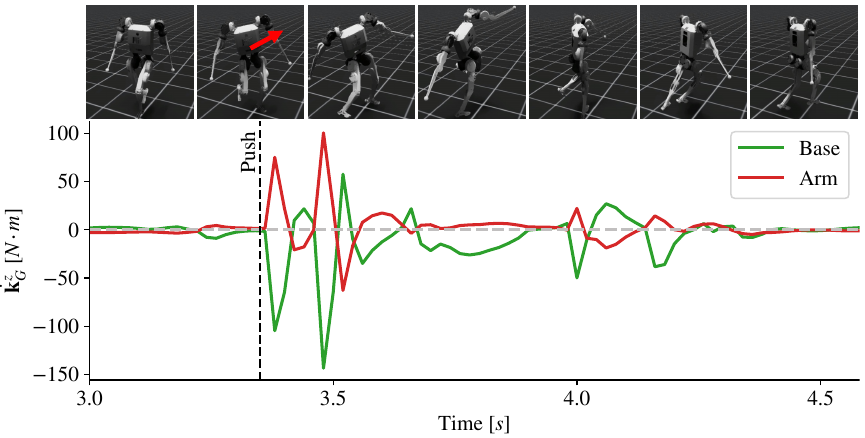}
    \caption{Emergent arm behavior under external perturbation. A simultaneous torque disturbance of 3 $N \cdot m$ along the $x$, $y$, $z$-axes is applied to the robot base at approximately 3.4 s (dashed line). The resulting abrupt time-rate of change of vertical CAM ($\dCAM^z$, green) in the base is quickly compensated by the emergent arm motion (red), producing opposite-direction momentum that stabilizes the overall CAM. The robot naturally recovers and returns to stable walking within approximately one second after the disturbance.}
    \label{fig:dCM_push}
\end{figure}

\begin{figure}[t]
    \centering
    \includegraphics[width=0.65\linewidth]{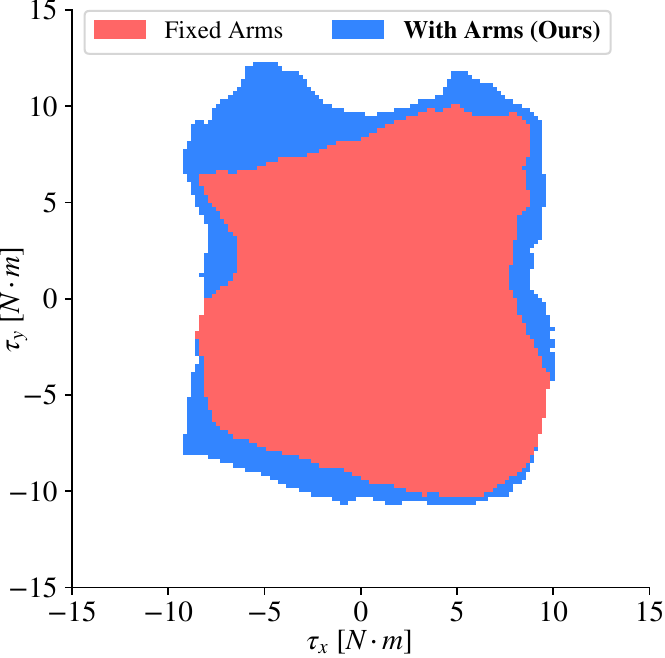}
    \caption{Comparison of disturbance recovery performance under random torque disturbances applied at $t = 1 s$, sampled from the range $[-15N \cdot m, 15N \cdot m]$ along the $x$, $y$-axes. The shaded regions represent successful recovery zones over the subsequent 4 seconds for the $\FixedArms$ (red) and our method (blue). The extended coverage of the blue region indicates that our controller remains balanced under a wider range of disturbances, approximately a 23\;\% improvement over the baseline range.}
    \label{fig:survival}
    \vspace{-3mm}
\end{figure}

Additionally, we investigate how the arm policy can assist with recovery from unexpected disturbances.
Such perturbations manifest as abrupt changes in the momentum rate of the robot's base, requiring rapid compensatory arm motion to stabilize the induced momentum.
\autoref{fig:dCM_push} demonstrates this emergent arm behavior under external disturbances.
In this experiment, we apply a 3 $N \cdot m$ torque disturbance to the base along the $x$, $y$, $z$-axes. 

The resulting jump in $\dCAM^z$ of the base induced by this perturbation is quickly countered by the emergent arm motion, which generates momentum in the opposite direction and effectively stabilizes the overall CAM.
Consequently, the robot swiftly recovers to a stable walking gait within approximately one second after the disturbance.

We further evaluate the robustness of our method by testing its ability to recover from a range of external disturbances.
Specifically, we apply random torque disturbances at time $t = 1 s$, sampled uniformly from the range $[-15N \cdot m, 15N \cdot m]$ along the $x$, $y$-axes, and evaluate the policy's ability to remain balanced over the subsequent 4 seconds.
As shown in \autoref{fig:survival}, our method achieves approximately 23\;\% higher success rate compared to the $\FixedArms$ baseline, demonstrating that the emergent arm motion contributes significantly to disturbance rejection, particularly under extreme perturbation conditions.

\subsection{Effectiveness of Limb-Level Multi-Agent RL}
\label{sec:marl_result}

\begin{figure}[t]
    \centering
    \includegraphics[width=0.99\linewidth]{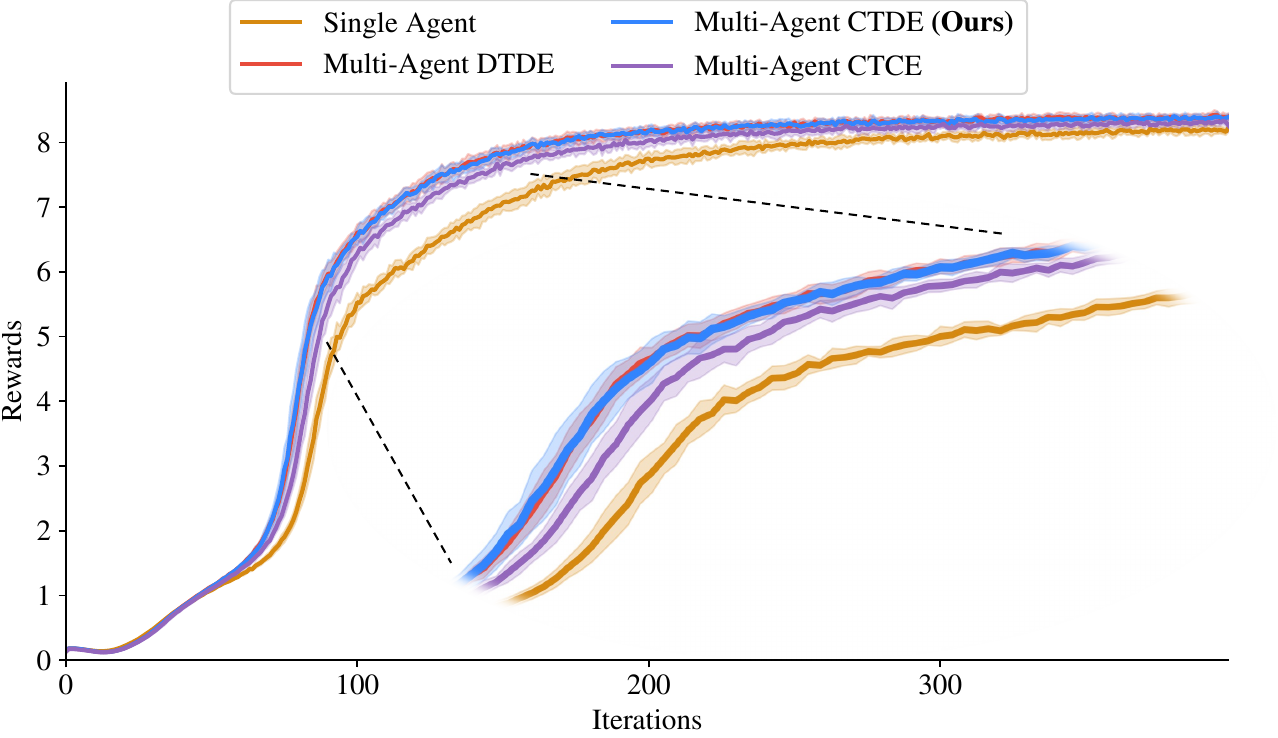}
    \caption{Reward convergence comparison of our method (\textit{Multi-Agent CTDE}) against three baselines over 400 training iterations. The plotted reward corresponds to the velocity tracking term defined in \autoref{eq:velocity_reward}. Both our approach and $\MultiAgentDTDE$ exhibit faster and higher convergence compared to the $\SingleAgent$ baseline and the $\MultiAgentCTCE$ variant. These results underscore the advantage of the CTDE design: using local observations for actors while sharing only base state and CAM enables stable and coordinated policy learning.
    }
    \label{fig:reward}
\end{figure}

In this section, we evaluate the effectiveness of our limb-level multi-agent RL framework, with an emphasis on the importance of the CTDE architecture.
To isolate the contribution of this framework, we compare our method against the following three baselines:

\begin{itemize}
\item \textbf{Single Agent}: A single actor-critic policy that receives the full global observation and outputs both arm and leg actions jointly.
\item \textbf{Multi-Agent DTDE}: A \textbf{D}ecentralized \textbf{T}raining and \textbf{D}ecentralized \textbf{E}xecution setup, where both actor and critic for each agent observe only local (limb-specific) and base information.
\item \textbf{Multi-Agent CTCE}: A \textbf{C}entralized \textbf{T}raining and \textbf{C}entralized \textbf{E}xecution setup, where both actor and critic for each agent observe the full global observation.
\end{itemize}

We begin by comparing the reward convergence of our approach (\textit{Multi-Agent CTDE}) against the three baselines over 400 training iterations, corresponding to approximately 8 minutes of wall-clock time, averaged over 10 different random seeds. 
The plotted reward corresponds to the velocity tracking term defined in \autoref{eq:velocity_reward}, which serves as the dominant objective for locomotion.
As shown in \autoref{fig:reward}, both our method and $\MultiAgentDTDE$ exhibit the fastest and highest reward convergence. 
In contrast, the $\SingleAgent$ baseline achieves significantly lower final reward, with noticeably slower convergence. The $\MultiAgentCTCE$ variant also performs slightly worse than our method.

In the $\SingleAgent$ setting, all arm and leg rewards are aggregated into a single scalar return, which can lead to conflicting objectives across limbs. 
As the policy attempts to optimize all reward terms simultaneously, balancing these objectives becomes a sensitive tuning problem, resulting in degraded performance across all components.
In contrast, our approach decouples the return for the arms and legs separately, allowing each policy to be updated without mutual interference.

In the case of $\MultiAgentCTCE$, each actor receives the full global observation, increasing the input dimension to 76.
This includes limb-specific information from the other agent, which is largely irrelevant and can introduce noise during policy learning. 
As a result, learning becomes less sample-efficient and convergence is hindered.

These findings support the CTDE design: actors should be restricted to local, limb-specific observations while sharing only base state and CAM, since CAM implicitly captures whole-body dynamics and provides sufficient latent information for coordination without unnecessary input noise.

\begin{figure}[tb]
    \centering
    \includegraphics[width=0.95\linewidth]{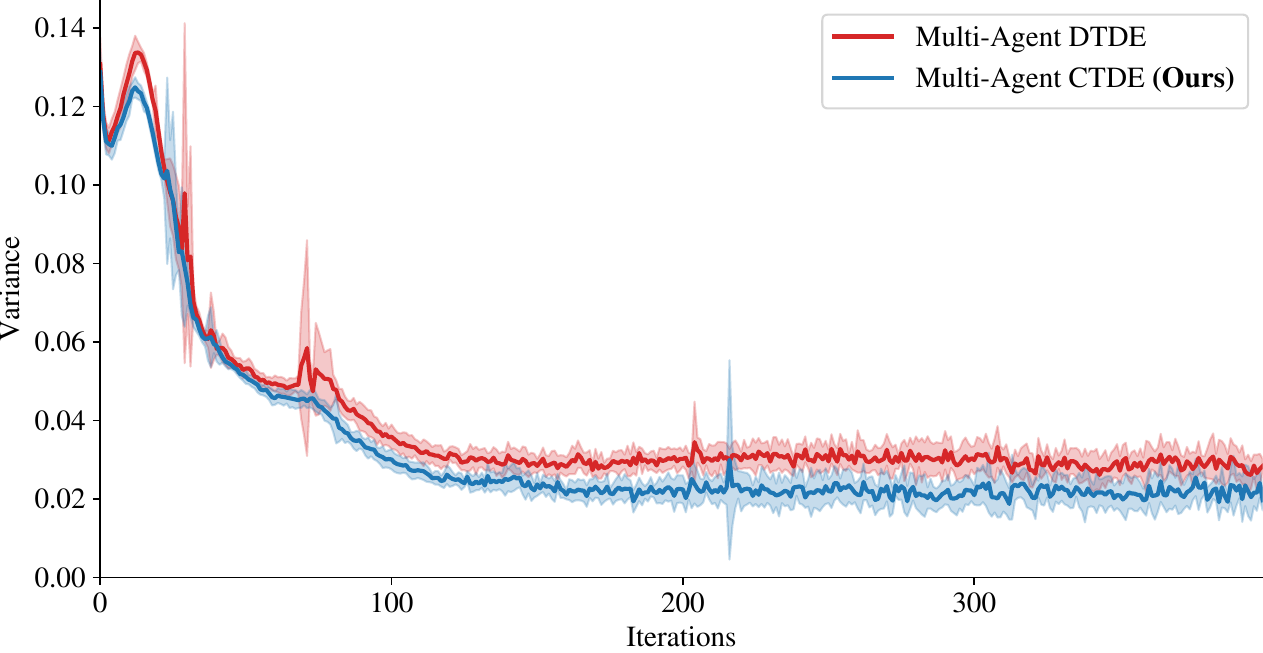}
    \caption{Variance of the advantage function for the arm agent during training, comparing $\MultiAgentDTDE$ and our method. Our approach exhibits faster convergence and consistently lower variance, indicating more stable and reliable advantage estimates. This supports the effectiveness of centralized critics, which leverage global observations to mitigate the multi-agent credit assignment problem and improve policy learning stability.}
    \label{fig:advantage}
\end{figure}

\begin{figure*}[tb]
    \centering
    \includegraphics[width=0.95\linewidth]{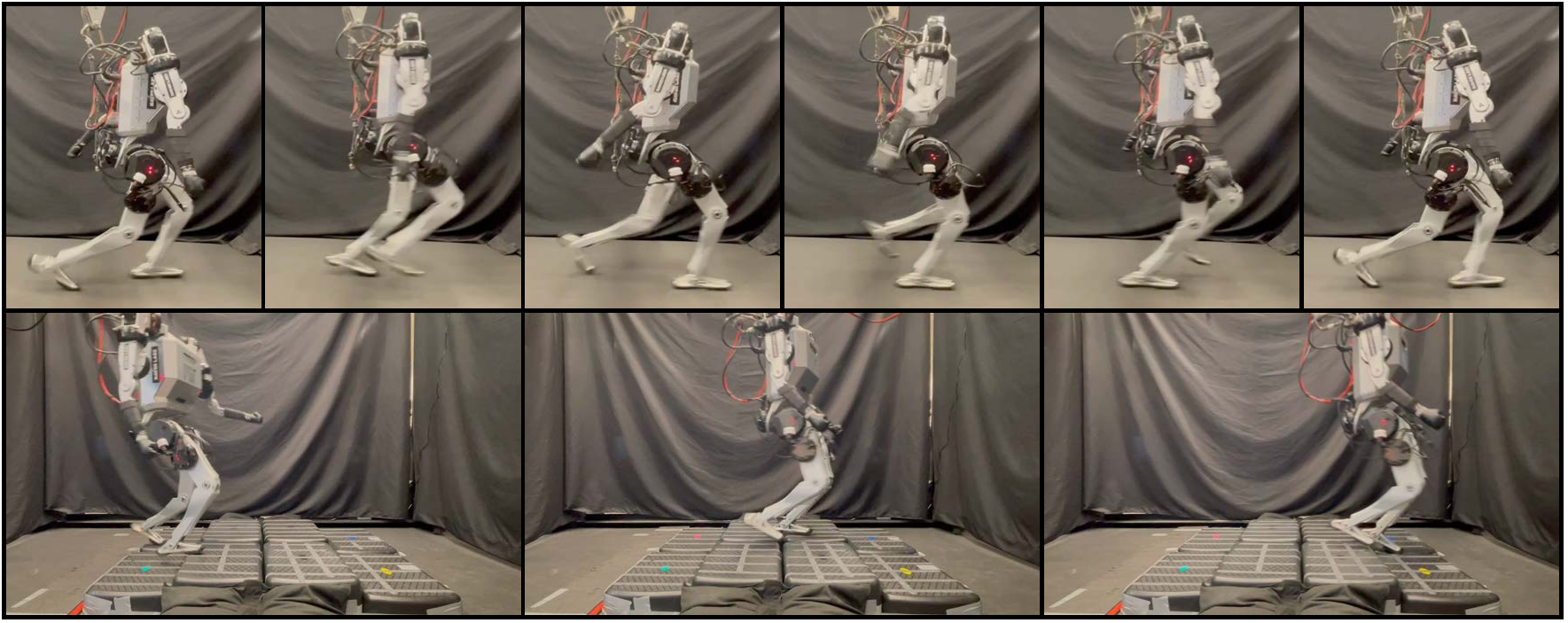}
    \caption{
     Hardware demonstrations of the learned policy on the \mithumanoid.
    \textit{Top Row:} Forward walking on flat ground with coordinated out-of-phase arm swing relative to the legs, reaching speeds up to 1.3 $m/s$. The sequence captures one complete gait cycle.
    \textit{Bottom Row:} Vision-free stair traversal, where adaptive arm motions assist in recovering from missteps and maintaining balance. See supplementary video for full motion sequences.
    }
    \label{fig:HW_test}
\end{figure*}

To better understand the distinction between $\MultiAgentDTDE$ and our approach (CTDE), we examine how policy gradients are estimated in RL. As shown in \autoref{eq:policy_gradient}, the policy update relies on the advantage function, which quantifies how good an action is relative to the expected value. 
This means that the consistency of the advantage estimate given a state is crucial for reliable policy gradient updates. 
High variance in the advantage function leads to noisy gradient estimates, potentially destabilizing learning. 
Conversely, lower variance suggests more stable and consistent learning signals. 
By measuring the variance of the advantage function throughout training, we can evaluate how effectively each framework mitigates the multi-agent credit assignment problem and promotes stable, consistent learning signals. 
To this end, we compare the advantage variance of the arm agent between $\MultiAgentDTDE$ and our approach in \autoref{fig:advantage}. 
Our method exhibits significantly lower final variance and converges more quickly, suggesting that providing the critic with global information is critical for producing reliable advantage estimates and facilitating more stable policy updates.

\subsection{Hardware Experiment}
\label{sec:hardware_experiment}

We demonstrate the real-world applicability of our approach by successfully transferring the learned policy to \mithumanoid~across a variety of locomotion tasks (\autoref{fig:HW_push}, \autoref{fig:HW_test}).
By extending the training environments from flat terrain to more diverse conditions—including rough surfaces, stairs, and inclined planes—we show that the emergent whole-body motion generalizes without modification. 
Our policy enables the robot to perform various tasks such as walking on flat ground, navigating uneven terrain, and ascending and descending stairs.

The top row of \autoref{fig:HW_test} illustrates forward walking with clear out-of-phase arm swing relative to the legs, achieving speeds up to 1.3 $m/s$. 
The bottom row shows the robot crossing stairs without vision, using adaptive arm motion to help recover from missteps and maintain balance. 
Detailed motion sequences can be found in the supplementary video.

%% file: Sections/06_conclusion.tex
\section{Conclusion}

In this work, we presented a multi-agent RL framework that leverages CAM objectives to learn the emergent arm swing and coordinated whole-body locomotion. 
To generate naturalistic arm motion while maintaining stable locomotion, we design separate reward structures for each agent: the leg agent is primarily guided by velocity tracking and contact rewards, while the arm agent is driven by CAM tracking and damping objectives. 
Our analysis shows that the learned arm behavior effectively cancels out the CAM generated by the rest of the body, reducing whole-body angular momentum. 
We further demonstrate that this CAM-driven arm motion lowers vertical GRM during walking and improves the system's ability to recover from external disturbances.
We also provide comparative evaluations showing that our multi-agent CTDE formulation outperforms alternative baselines, including single-agent and centralized actor configurations. 
These findings underscore the benefit of modular limb-level policies for stable, coordinated control. 
Finally, we validate the practicality of our approach on \mithumanoid, demonstrating successful transfer to diverse real-world locomotion tasks such as walking on flat ground, navigating rough terrain, and climbing stairs.

In future work, we plan to explore more dynamic and contact-intensive scenarios, including one-foot balancing, running, and complex tasks such as dribbling or kicking. 
We believe that combining our control framework with human motion datasets will further enhance the naturalness and transferability of learned behaviors to real humanoid robots.